\documentclass[acmtog,nonacm]{acmart}

\usepackage{booktabs} 

\usepackage{mathtools}
\usepackage{multirow}

\usepackage[ruled]{algorithm2e} 

\SetAlFnt{\small}
\SetAlCapFnt{\small}
\SetAlCapNameFnt{\small}
\SetAlCapHSkip{0pt}

\acmJournal{TOG}

\begin{document}

\title{Learning Implicit Glyph Shape Representation}

\author{Ying-Tian Liu}
\affiliation{%
 \institution{Tsinghua University}
 \department{BNRist, Department of Computer Science and Technology}
 \city{Beijing}
 \country{China}
}
\email{liuyingt20@mails.tsinghua.edu.cn}

\author{Yuan-Chen Guo}
\affiliation{%
 \institution{Tsinghua University}
 \department{BNRist, Department of Computer Science and Technology}
 \city{Beijing}
 \country{China}}
\email{guoyc19@mails.tsinghua.edu.cn}

\author{Yi-Xiao Li}
\affiliation{%
 \institution{Tsinghua University}
 \department{BNRist, Department of Computer Science and Technology}
 \city{Beijing}
 \country{China}}
\email{liyixiao20@mails.tsinghua.edu.cn}

\author{Chen Wang}
\affiliation{%
 \institution{Tsinghua University}
 \department{BNRist, Department of Computer Science and Technology}
 \city{Beijing}
 \country{China}}
\email{wchen20@mails.tsinghua.edu.cn}

\author{Song-Hai Zhang}
\affiliation{%
 \institution{Tsinghua University}
 \department{BNRist, Department of Computer Science and Technology}
 \city{Beijing}
 \country{China}}
\email{shz@tsinghua.edu.cn}

\begin{abstract}

Automatic generation of fonts can greatly facilitate the font design process, and provide prototypes where designers can draw inspiration from. Existing generation methods are mainly built upon rasterized glyph images to utilize the successful convolutional architecture, but ignore the vector nature of glyph shapes. We present an implicit representation, modeling each glyph as shape primitives enclosed by several quadratic curves. This structured implicit representation is shown to be better suited for glyph modeling, and enables rendering glyph images at arbitrary high resolutions. Our representation gives high-quality glyph reconstruction and interpolation results, and performs well on the challenging one-shot font style transfer task comparing to other alternatives both qualitatively and quantitatively.

\end{abstract}

%
%


%
%


\begin{teaserfigure}
\centering
\includegraphics[width=\linewidth]{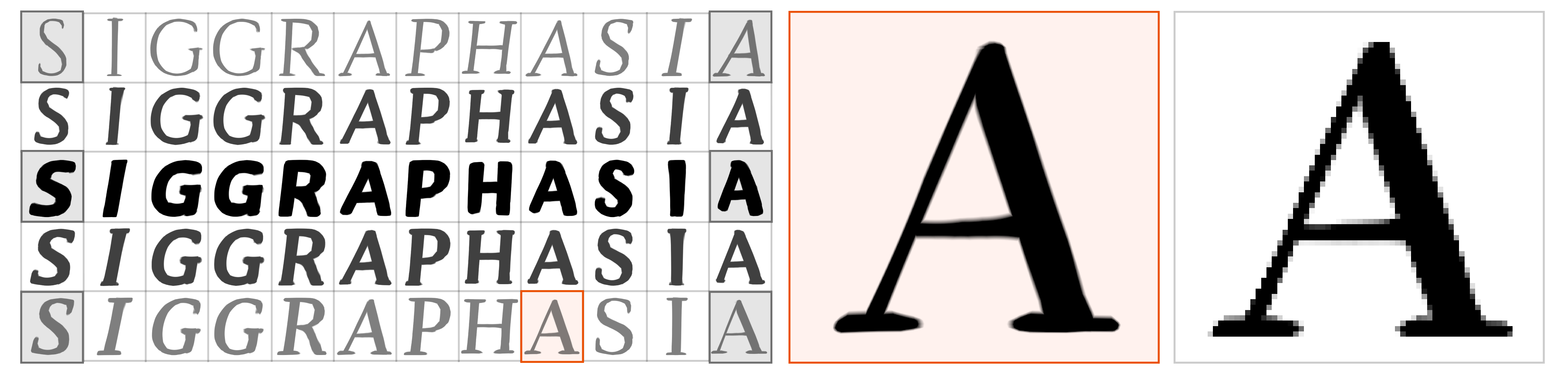}
\caption{Left: Given glyphs on the gray background, we linear interpolate their styles to generate glyphs in new styles, using our implicit glyph shape representation. Each glyph is in a different style. Right: On the red background is a high-resolution (2048$\times$2048) glyph image produced by our proposed representation, compared with an aliased version generated by magnifying a low-resolution (64$\times$64) glyph image.}
\label{fig:teaser}
\end{teaserfigure}

\maketitle
\thispagestyle{empty}

\section{Introduction}
\label{introduction}

Font design is a labor-intensive task. It could take days, or even years, for professional artists to design a new font, considering the size of character set and the complexity of the font style. Automatic font generation aims to facilitate this tedious process, by providing a set of style-coherent glyphs from a small amount of manually-designed samples~\cite{example-based, dcfont, font-cgan}. With the success of deep learning techniques in natural image synthesis, recent works mainly adopt a similar architecture, which try to solve an image synthesis problem by regarding glyph shapes as rasterized images~\cite{agisnet, mcgan, glyphgan}. However, glyph shapes are inherently vector graphic elements describing how the shapes are drawn, thus supporting rendering at arbitrary resolutions. This gap between vector fonts and rasterized images raises two main problems for the image synthesis approaches: first, no structural information is explicitly modeled through convolutions on rasterized images, which is fundamental for building cross-character dependencies; second, the output image resolution is limited, and it is not straight-forward to convert the glyph image to a vector representation, which significantly limits the applicability of these methods.

This is much like the case when using voxels to represent 3D shapes. Voxel naturally fits into the convolutional architecture, but lacks structural information and brings the problem of high computational costs and memory inefficiency as its size grows. To overcome these drawbacks, in recent years, implicit functions are widely used to represent 3D shapes, which model the occupancy value or distance to the nearest surface for any continuous 3D location. Based on the implicit representation, methods like BSP-Net~\cite{bspnet} and CvxNet~\cite{cvxnet} model 3D shapes as unions of convexes constructed from hyperplanes, recovering clear shape boundaries.

Taking inspirations from 3D shape modeling, we propose to represent glyph shapes as primitives enclosed by quadratic curves. We find this kind of implicit representation well suited for glyph shape modeling for two reasons. First, shapes are constructed from parts described by parametric lines in a hierarchical manner, which highly resembles the actual design process. Second, the ability to evaluate at continuous locations allows rendering at arbitrary resolutions, and it is feasible to convert the representation to common vector formats like SVG and TTF, for practical use.

Our representation for glyph shapes starts from groups of quadratic curves. Each group of curve formulates a shape primitive, by max operations on their signed distance functions. Shape primitives are further combined to be a glyph shape, by min operations on their signed distance functions. We show that parameters of the quadratic curves can be learned from glyph image features, resulting in better reconstruction quality comparing to existing CNN-based methods. Our representation is explainable by visualizing curves and shape primitives, which are shown to be highly consistent for different fonts of the same character. We also demonstrate the power of this representation on a challenging one-shot font style transfer task, where only a single glyph image is presented to get a full set of style-coherent glyphs. Qualitative and quantitative results support the effectiveness of this representation in formulating glyph shapes. 

Our contributions are threefold:
\begin{itemize}
    \item A novel implicit representation for glyph shapes is proposed, which models glyph shapes as combination of primitives enclosed by quadratic curves, enabling glyph image rendering at arbitrary resolutions for practical use.
    
    \item Experiments show that our model produces realistic glyph reconstruction and interpolation results with clear boundaries and clear structures, suggesting that a good latent manifold of glyphs can be learned without extra elaborate network designs.
    
    \item Both qualitative and quantitative experiments on the challenging one-shot font style transfer task verify the superiority of our method over other alternatives, further indicating the effectiveness of the proposed representation. 
\end{itemize}
\section{Related Work}
\label{related-work}
\subsection{Font Generation}
Font generation aims at simplifying the workflow of manually designing new fonts. Existing methods on font generation typically work under two scenarios. The first one aims to help professional designers on the font design process. Existing works mainly take a few glyph images of the same style as references and generate the other glyphs of this style, which is also referred as font style transfer.
The second one benefits both novice users and experts by generating an entire new font set via style interpolation of existing fonts or manifold learning. In terms of the shape representation of generated glyphs, these methods can be further classified into two categories: vector font generation and glyph image generation. 

\subsubsection{Vector Font Generation}
Generating vector fonts is a classical problem in the area of computer graphics. Some works~\cite{campbell2014, learn_manifold} proposed to learn a font manifold and produced a generative model enabling users to manipulate the style of the entire font set by modifying a single glyph.
Lian et al.~\cite{lian2018} proposed a system to automatically generate large-scale handwriting fonts by utilizing automatic stroke extraction and handwriting detail recovery techniques. Gao et al.~\cite{gao19} proposed a Chinese vector fonts generation system by inferring layout information and assembling corresponding components. Most related to this work, Lopes et al.~\cite{lopes19} proposed to learn latent representations for the Scalable Vector Graphics (SVG) format, by sequential generative modeling. Comparing to our implicit representation, it is not easy to model SVG format in a deep learning architecture, and not straight-forward to render SVG elements to images in a differentiable manner, thus requiring ground-truth SVG drawing commands. In contrast, our representation is based on signed distance functions, which is easy to model with neural networks and can be learned without additional supervisions other than input images.

\subsubsection{Glyph Image Generation}
Recently, deep convolutional neural networks have attracted a lot of interests in glyph image synthesis. With the great success of Variational Auto-encoders (VAEs)~\cite{vae14} and Generative Adversarial Networks (GANs)~\cite{goodfellow14}, many works have been proposed to transfer styles of glyph shapes~\cite{glyphgan} or glyph texture effects~\cite{tetgan} using these techniques. Isola et al.~\cite{pix2pix} proposed \textit{pix2pix}, a general image-to-image translation framework, that can be used for various image generation tasks. Based on \textit{pix2pix}, 
Azadi et al.~\cite{mcgan} proposed an end-to-end solution (MC-GAN) to synthesise ornamented glyphs for 26 Latin capital letters from a small number of glyph observations of the same style. AGIS-Net~\cite{agisnet} proposed by Gao et al. transferred both shape and texture styles to arbitrarily large numbers of characters with a few references. STEFANN~\cite{stefann} achieved scene text editing via structure and color transfer in a two-stage manner. In addition, Wang et al.~\cite{attribute2font} studied controllable font generation from descriptive attributes, allowing novice users to create their own fonts from intuitive thoughts. However, image-based methods are more likely to suffer from blurring and ghosting artifacts since structured information of glyphs are typically lost during rasterization, and cannot scale to arbitrary resolutions. Our representation models glyphs with primitives enclosed by quadratic curves, which ensures clear boundaries, learns hierarchical structures while supporting rendering at any resolution. 

\subsection{Implicit Shape Representation}
Implicit functions like Occupancy Functions and Signed Distance Functions (SDFs) have wide applications in 3D shape modeling and rendering. As deep learning becomes a trend in the community, deep implicit functions (DIFs), which are implicit functions parameterized by neural networks, are becoming more and more popular due to their compactness and strong representation power~\cite{implicitfield, occupancy_net, deepsdf}. Chen et al.~\cite{implicitfield} for the first time introduced a deep network for learning implicit fields for generative shape modeling. Later, Chen et al. ~\cite{baenet} extended the work to a branched 3D auto-encoder network to perform co-segmentation for 3D shapes in an unsupervised way. Based on DIFs, BSP-Net~\cite{bspnet} and Cvx-Net~\cite{cvxnet} modeled shapes as a combination of convex sets. We believe that this representation is also suitable for the representation of 2D graphics such as glyphs, and is superior to convolution-based representation since glyphs have clear boundaries and can be easily split into parts. We extend this representation by adopting quadratic curves as basic elements for glyph shape modeling, and achieve state-of-the-art results on various tasks.

\textbf{}\section{Methodology}
\subsection{Implicit Glyph Shape Representation}

\begin{figure}[htbp]
    \centering
    \includegraphics[width=\linewidth]{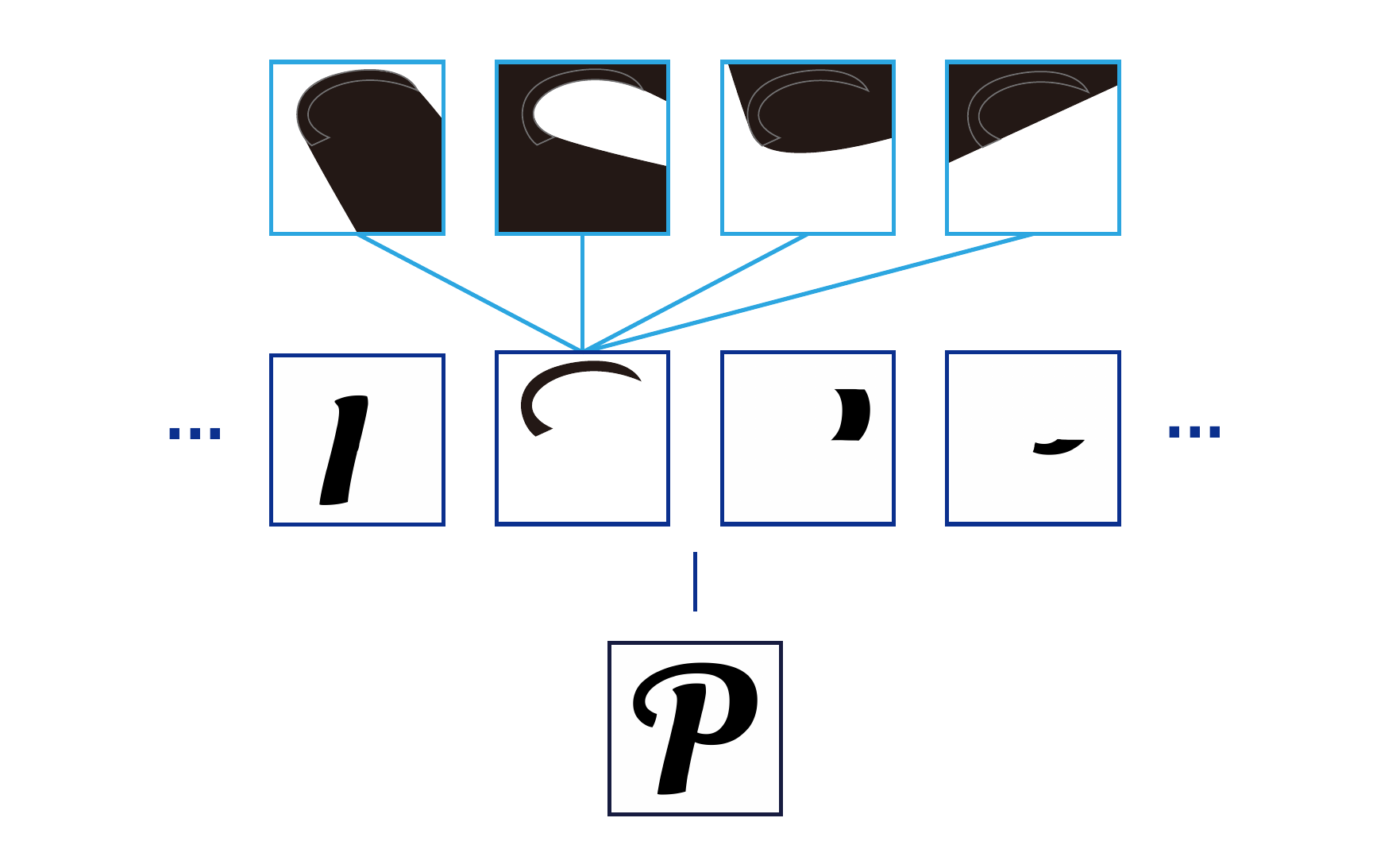}
    \caption{Illustration of the implicit glyph shape representation. Black area is the interior of the shape. Glyphs (third row) are composed of primitives (second row) enclosed by quadratic curves (first row).}
    \label{fig:illustration}
\end{figure} 

We first formulate our implicit representation for modeling glyph shapes. We model a 2D shape as a signed distance function $SDF$ defined on continuous 2D coordinates $\mathbf{x}\in\mathbb{R}^2$:
\begin{equation}
    SDF(\mathbf{x}) = s \in \mathbb{R} \left\{
\begin{aligned}
\le 0 & , inside \\
 > 0  & , outside \\
\end{aligned}
\right.
\end{equation}
where the absolute value of $s$ indicates the distance from $\mathbf{x}$ to its closest boundary. $\mathbf{x}$ is inside the shape when $s\le0$ and outside otherwise. Zero iso-surface of the function $SDF(\mathbf{x}) = 0$ represents the boundary of the shape. While it is possible to model $SDF$ with MLPs by taking a 2D coordinate as input and directly giving $s$, we take an explainable way by constructing the signed distance field from curves in 2D space. The details are explained mathematically as follows.

Let there be $vp$ quadratic curves, divided into $v$ groups, each containing $p$ curves. Let $P\in\mathbb{R}^{v\times p\times 6}$ be the parameters of these quadratic curves. For an arbitrary 2D location $\mathbf{x}=(x,y)$, the signed distance of $\mathbf{x}$ with respect to the $j$th curve in the $i$th group, or $d_{ij}(\mathbf{x})$, is:
\begin{equation}
\label{equation:dij}
    d_{ij}(\mathbf{x}) = [x^2, xy, y^2, x, y, 1]P_{ij}^T
\end{equation}
where $P_{ij}\in\mathbb{R}^6$ is the parameter of this curve.

\begin{figure}[!t]
    \centering
    \includegraphics[width=\linewidth]{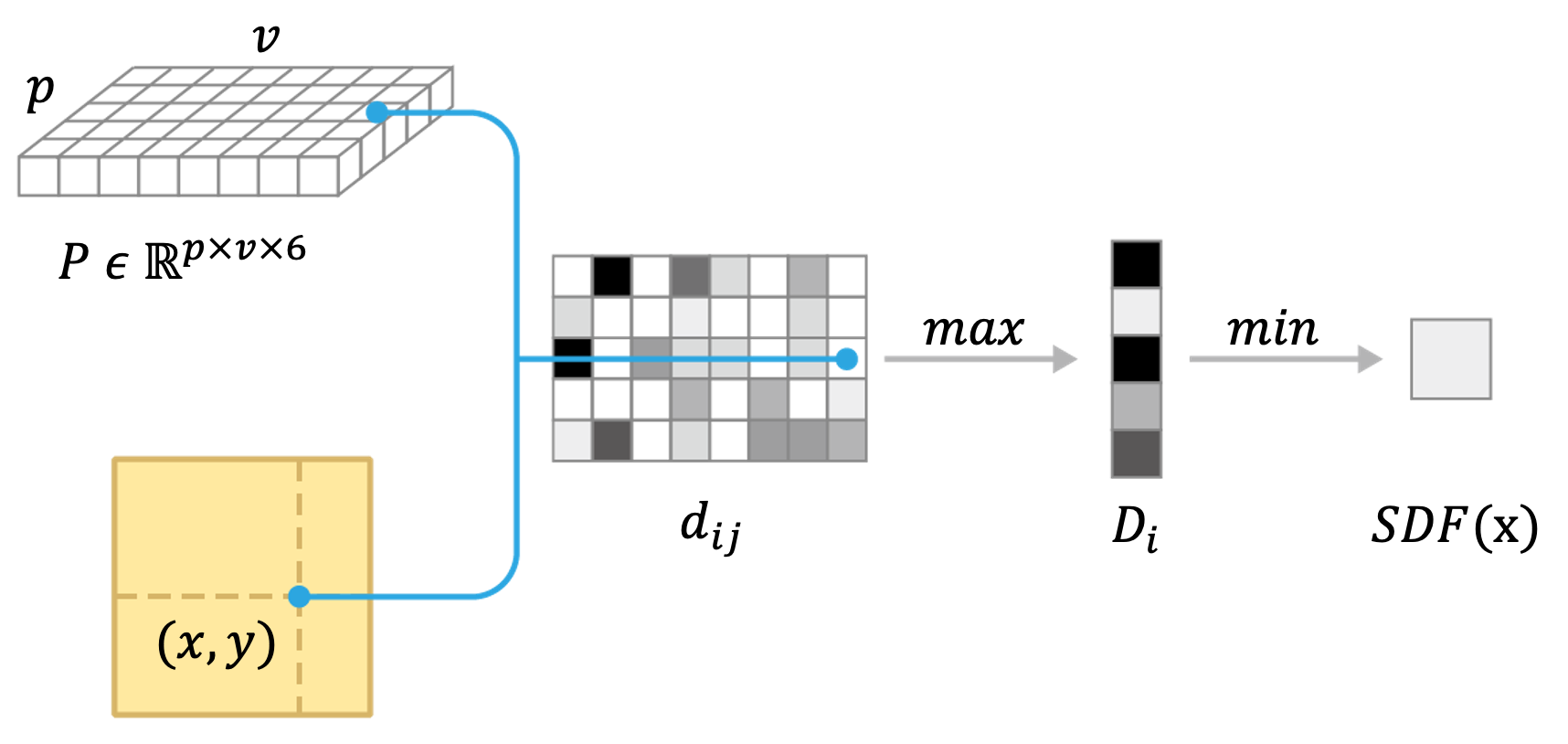}
    \caption{The calculation process of obtaining signed distance value for a 2D position $(x,y)$. We refer to this process as ``implicit decoder''.}
    \label{fig:implicit_decoder}
\end{figure} 

By applying a max operation on the signed distances of each group $\{d_{i(\cdot)}\}$, we get $D_i(\mathbf{x})$, which represents a primitive enclosed by curves of this group:
\begin{equation}
\label{equation:di}
    D_{i}(\mathbf{x}) = \max_j d_{ij}(\mathbf{x}) \left\{
\begin{aligned}
\le 0 & , inside \\
 > 0  & , outside \\
\end{aligned}
\right.
\end{equation}

Finally, the shape is represented as the union of all primitives, by a min operation on all $D_{i}(\mathbf{x})$:
\begin{equation}
\label{equation:sdf}
    SDF(\mathbf{x}) = \min_i D_i(\mathbf{x}) \left\{
\begin{aligned}
 \le 0 & , inside \\
 > 0  & , outside \\
\end{aligned}
\right.
\end{equation}

Fig.\ref{fig:illustration} is an intuitive illustration of using this representation for modeling glyph shapes. The black area is the interior of the shape, where $SDF(\mathbf{x})\le 0$. Quadratic curves (first row) form primitives (second row), which construct the final glyph shape (third row). Fig.\ref{fig:implicit_decoder} shows how the signed distance value is obtained from curve parameters $P$. We term this process ``implcit decoder" in the remainder of the text.

\subsection{Representation Learning}
\label{subsec:network-architecture}
To learn such an implicit representation for glyph images, we adopt an encoder-decoder architecture. A CNN-based encoder $\mathcal{E}_I$ extracts features from glyph images, which are fed to an MLP $M_{param}$ to predict curve parameters $P$. Signed distance value for each pixel location $\mathbf{x}$ can then be calculated according to
Eq.\ref{equation:dij},\ref{equation:di},\ref{equation:sdf}.

We replace the max operation with summation to ease training, while adopting a learnable coefficient $\sigma_{v \times p}$ to preserve the magnitude of $D^+_{i}(\mathbf{x})$. We fix the number of curves in each primitive, making the training process more stable. 
\begin{equation}
    D^+_{i}(\mathbf{x}) = \sum_j ReLU(\sigma_{ij} d_{ij}(\mathbf{x})) \left\{
\begin{aligned}
 = 0 & , inside \\
 > 0  & , outside \\
\end{aligned}
\right.
\end{equation}

For the min operation in Eq.\ref{equation:sdf}, we replace it with summation weighted by a learnable vector $W$:
\begin{equation}
SDF^{+}(\mathbf{x})=1 - \left[\sum_{i} W_{i}\left[1-D_{i}^{+}(\mathbf{x})\right]\right]_{[0,1]}\left\{\begin{array}{l}
= 0 \approx inside \\
{(0,1] \approx outside}
\end{array}\right.
\label{equ:approx-sum}
\end{equation}

\begin{figure*}[htbp]
    \centering
    \includegraphics[width=\linewidth]{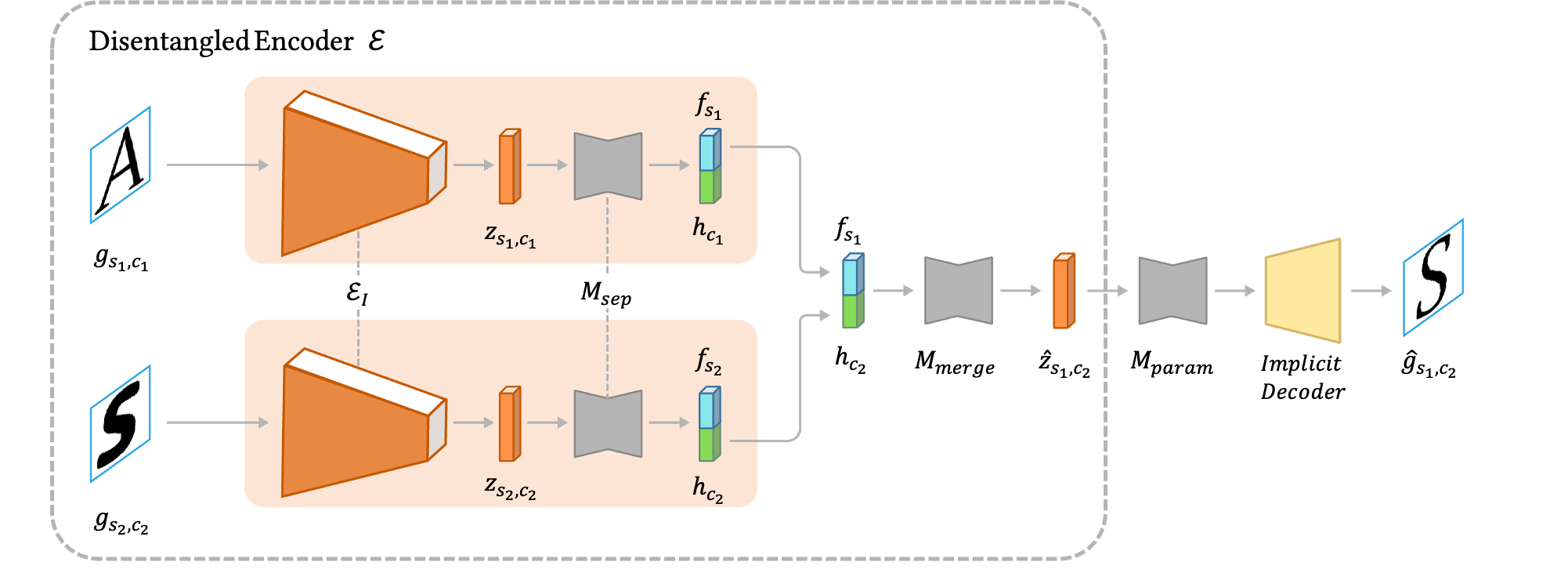}
    \caption{Illustration of the network architecture for one-shot font style transfer. Style feature $f_{s_1}$ from the upper input image $g_{s_1,c_1}$ and content feature $h_{c_2}$ from the lower input image $g_{s_2,c_2}$ are extracted respectively and merged to get a new feature vector $\hat{z}_{s_1,c_2}$. It is then used to predict curve parameters $P$, and produce the output glyph by our implicit decoder. The output glyph $\hat{g}_{s_1,c_2}$ should have the content of $g_{s_2,c_2}$ and share the same style with $g_{s_1,c_1}$.}
    \label{fig:disentangle_arch}
\end{figure*} 

Note that $SDF^{+}(\mathbf{x})$ is no longer a signed distance value, but an approximated occupancy value. To reconstruct the input glyph image, we compare $SDF^{+}(\mathbf{x})$ with the pixel value at each location $\mathbf{x}$ by $L_2$ distance:
\begin{equation}
    L_{rec} = \mathbb{E}_{\mathbf{x}\sim T}[\lVert SDF^{+}(\mathbf{x}) - G(\mathbf{x}|g) \rVert_2]
\end{equation}
where $T$ contains sampled pixel locations, normalized to $[-1,1]$. $G(\mathbf{x}|g)$ denotes the pixel value of glyph image $g$ at location $\mathbf{x}$. Also, we would like the weight vector $W$ in Eq.\ref{equ:approx-sum} to be close to $1$ so that the it is a good approximation of the min operation:
\begin{equation}
    L_{W} = \sum_i |W_i - 1|
\end{equation}

The overall optimization objective is the weighted average of the two:
$$L = L_{rec} + \lambda_{W}L_{W}$$

\section{One-Shot Font Style Transfer}
\label{network}

\subsection{Problem Definition}
To further demonstrate the potential for font generation of our proposed representation, we bring up a challenging one-shot font style transfer problem, which is formulated as follows. 

Denote a font set $S$ to be a set of $N$ \textit{glyphs} $\{g_{s, 1}, g_{s, 2}, ... g_{s, N}\}$. In this paper we only consider the uppercase Latin characters ($N=26$), but it is natural to extend to other character sets. All glyphs from the same font set should share a common style, denoted as the first subscript $s$, but have different contents, denoted as the second subscript $i$. Given a specific glyph $g_{s, c_1}$ with arbitrary content as \textit{the style reference}, the goal is to generate all other glyphs in the font set.

To better extend to larger font sets, we translate the problem into a single glyph generation problem with a style and a content reference. Given the style reference $g_{s_1, c_1}$ and content reference $g_{s_2, c_2}$, we aim to generate $g_{s_1, c_2}$ with style $s_1$ and content $c_2$. To generate the whole font set, we traverse the entire character set and take each glyph as the content reference. Note that, ideally, the style of the content reference and the content of the style reference will not affect the generation result.

To solve this problem, we propose a model consisting of a disentangled encoder $\mathcal{E}$ and an implicit decoder, as illustrated in Fig.~\ref{fig:disentangle_arch}. Details are described as follows.

\subsection{Network Architecture}
\subsubsection{Disentangled Encoder}
The disentangled encoder $\mathcal{E}$ consists of a convolutional neural network $\mathcal{E}_I$ to get latent feature vectors for input images, and two MLPs $M_{sep},M_{merge}$ to separate the features into content and style features and re-merge them.

More specifically, the CNN encoder $\mathcal{E}_I$ takes $g_{s_1, c_1}, g_{s_2, c_2}$ as inputs and encodes them into latent vectors $z_{s_1, c_1}, z_{s_2, c_2}$. After that, these vectors are fed to the separate network $M_{sep}$, to be split into style vectors $f_{s_1}, f_{s_2}$ and content vectors $h_{c_1}, h_{c_2}$. We then concatenate $f_{s_1},h_{c_2}$ and feed them into the merge network, $M_{merge}$, to produce $\hat{z}_{s_1, c_2}$, which approximates the feature vector of $g_{s_1,c_2}$. It is further used to estimate curve parameters by $M_{param}$ and output the desired glyph by our implicit decoder.

\subsubsection{Auxiliary Character Classifier}
While the above process is already able to accomplish the task of font style transfer, we observed that when the content reference $g_{s_2, c_2}$ has complex style whose character is hard to be recognized, the generated $\hat{g}_{s_1, c_2}$ often carries ambiguous content information. We therefore design an auxiliary character classifier supervised on the category of the character generated. The classifier $\mathcal{C}$ is a convolutional network and predicts what the character is for the input image. We address that the auxiliary character classifier is essential for generating correct characters, by providing explicit guidance on image contents. 

\subsection{Optimization Objectives}
In this section, we describe the optimization objectives used for training our one-shot font style transfer model.

In order to effectively disentangle the content and style information, we encourage the content vectors obtained from $g_{s_2, c_2}$ and $g_{s_1, c_2}$ to be the same via a content loss:
\begin{equation}
    L_{cont} = \mathbb{E}[|M_{sep}(z_{s_1, c_2})_h - M_{sep}(z_{s_2, c_2})_h|]
\end{equation}
where the subscript $h$ indicates that we only calculate the differences between content vectors. The same routine is applied for ensure style consistency:
\begin{equation}
    L_{style} = \mathbb{E}[|  M_{sep}(z_{s_1, c_1})_f - M_{sep}(z_{s_1, c_2})_f |]
\end{equation}
From definition, $M_{sep}$ and $M_{merge}$ are inverse process of each other. We calculate a latent vector reconstruction loss on the latent vector $z_{s_1, c_2}$ fed into $M_{sep}$ and $\hat{z}_{s_1, c_2}$ produced by $M_{merge}$:
\begin{equation}
    L_{latent} = \mathbb{E}[| z_{s_1, c_2} - \hat{z}_{s_1, c_2} |]
\end{equation}
Since the goal of $\mathcal{C}$ is to recognize the content from a given image, we apply a cross-entropy loss:
\begin{equation}
    L_{c} = \mathbb{E}[CE(\mathcal{C}(g_{s, c}), c)]
    \label{loss-classifier}
\end{equation}
Then we have the following category loss on the generated image as follows:
\begin{equation}
    L_{cate} = \mathbb{E}[CE(\mathcal{C}(\hat{g}_{s_1, c_2}), c_2)]
\end{equation}
Note that the only difference between the calculation of $L_{cate}$ and $L_{c}$ is that $L_{cate}$ depends on the generated images while $L_{c}$ learns from ground truth inputs.

During training, we optimize the model parameters by the following objectives:
\begin{equation}
\begin{aligned}
  L = & L_{rec} + \lambda_W L_W + \lambda_{cont} L_{cont} +  \lambda_{style} L_{style} + \\ &\lambda_{latent} L_{latent} + \lambda_{cate} L_{cate}
\end{aligned}
\end{equation}
where $L_{rec}$ and $L_W$ are objectives described in Sec.\ref{subsec:network-architecture}.

\section{Experiments}
\label{experiments}
We conduct extensive experiments to show the nice properties of our representation in glyph modeling, and its great potential in font generation. We first illustrate that our representation can not only reconstruct glyphs with rich details and clear boundaries, but also give promising results in interpolating glyphs of different styles. Quantatitive and qualitative comparisons on the one-shot font style transfer task further support our claim. We also perform ablation studies to show the effectiveness of several design choices, and discuss some typical failure cases to inspire future directions.
\subsection{Experiment Settings}
\subsubsection{Implementation details}
In our experiments, all input images are in size 64 $\times$ 64. We use ResNet-18~\cite{resnet} with Leaky ReLU~\cite{lrelu} and Batch Normalization~\cite{bn} as image feature extractor for $\mathcal{E}_I$ and $\mathcal{C}$. For the implicit decoder, we have $v = 16$ primitives and $p = 6$ curves for each primitive. $M_{param}$ consists of two fully connected layers of size 256 and outputs $6vp$ curve parameters. In the one-shot font style transfer problem, dims for latent vector $z$, style vector $f$ and content vector $h$ are $256$, $128$ and $128$. $M_{sep}$ and $M_{merge}$ consist of $2$ and $3$ fully connected layers of size $256$ respectively. Weights in the loss function are selected as 
$\lambda_W = 0.1, \lambda_{cont} = 0.1, \lambda_{style} = 0.1, \lambda_{latent} = 0.1$ and $\lambda_{cate} = 0.05$. We use Adam optimizer~\cite{adam} with $lr = 1e-4$ and $betas=(0,9, 0.999)$. We train the reconstruction model with batch size 32 for 60,000 iterations, and train the one-shot generation model with batch size 32 for 200,000 iterations.

\subsubsection{Dataset}
\begin{figure}[htbp]
    \centering
    \includegraphics[width=\linewidth]{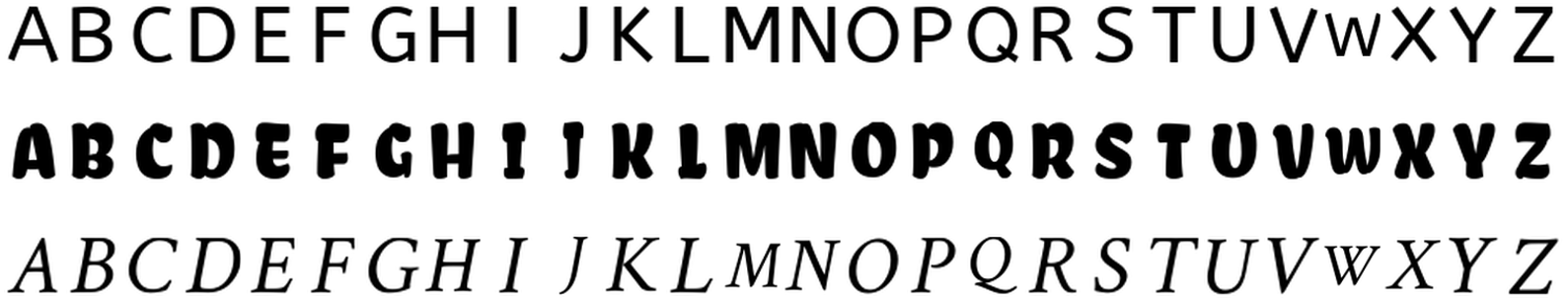}
    \caption{Examples of glyph images from the dataset we used for training and evaluation.}
    \label{font-example}
\end{figure}

We train and evaluate all our models on the dataset provided by ~\cite{stefann}. It contains 1315 fonts, 1015 of which are split for training the rest 300 for validation. Each font has 26 uppercase Latin letters, represented as 64 $\times$ 64 glyph images. Example glyph images from the dataset are shown in Fig.\ref{font-example}.
We select glyphs from \textit{abeezee-regular} (top row of Fig.~\ref{font-example}) as the content references for those methods which need fixed content references in training.
For fair comparison, we use glyphs from this font as the content references for all methods in evaluation.

\subsubsection{Compared Methods}
To demonstrate the superiority of our proposed representation on glyph modeling, we compare our method with four baseline approaches, including general-purpose image generation models (VAE, pix2pix), and state-of-the-art glyph image generation models (AGIS-Net, FANnet). We briefly introduce the four baselines as well as the training settings as follows.
\begin{itemize}
    \item Variational Autoencoder (VAE)~\cite{vae14}: VAE enables reconstruction of data while maintaining a latent distribution which could be sampled from to generate new data. We compare with it on glyph reconstruction and interpolation.
    \item pix2pix~\cite{pix2pix}: pix2pix is a popular general-purpose image-to-image translation model. It learns a conditional mapping from the source image domain to the target image domain. It adopts a U-Net~\cite{unet} architecture and trains in an adversarial manner. We compare with pix2pix on the one-shot font style transfer task. To provide pix2pix with both style and content information, we simply stack the two images at the channel dimension, and use the stacked tensor as input. 
    \item AGIS-Net~\cite{agisnet}: AGIS-Net is the state-of-the-art artistic glyph image generation model. It uses two encoders to extract style and content information separately and two decoders for gray-scale and textured glyph image generation. It adopts shape, texture and local discriminator, and trains adversarially. It transfers both the shape and texture styles under few-shot observations and provides promising results. We only train AGIS-Net to transfer shape styles under a single observation and compare with it on the one-shot font style transfer task. 
    \item FANnet~\cite{stefann}: FANnet is the state-of-the-art text editing model. It follows a simple encoder-decoder architecture, which inputs a one-hot content code with a glyph image as style reference, and outputs a style-coherent glyph image in that content. It naturally supports one-shot generation, and we compare with it on the one-shot font style transfer task.
\end{itemize}

\subsubsection{Evaluation Metrics}
For quantitative evaluation, we apply several commonly-used metrics for image generation: $L_1$ distance, structural similarity (SSIM)~\cite{ssim}, learned perceptual image patch similarity (LPIPS)~\cite{lpips}. SSIM can measure the structural similarity between the generated glyph images and ground truth images, which is more suitable for the font shape style evaluation. LPIPS is a learned human perceptual similarity metric, which prefers images with similar textures and is designed and trained on natural images. For all the three metrics, lower is better. 

\subsection{Representational Ability}
\subsubsection{Glyph Reconstruction}
\label{sec:reconstruction}
\begin{figure}[htbp]
    \centering
    \includegraphics[width=0.8\linewidth]{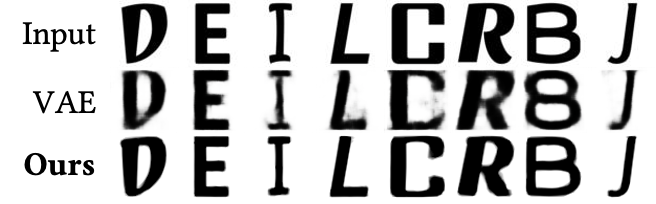}
    \caption{Visual comparisons of glyph images reconstructed by VAE~\cite{vae14} and our model using the implicit representation. Our results maintain clear boundaries and rich details, while showing promising generalizability on out-of-distribution data.}
    \label{fig:reconstruction}
\end{figure}

How well a model can reconstruct its inputs gives an impression on the representational power of the model. In this experiment, we train our model to reconstruct the input glyph images. A total of $26390$ glyph images from the training font set are used for training, and comparisons are conducted on the validation set. The same setting is applied on VAE~\cite{vae14} for fair comparison. We evaluate SSIM between reconstructed and original images for quantitative comparison. Our model achieves an SSIM score of $0.9039$, which is significantly higher than $0.7752$ for the VAE model. Qualitative results are shown in Fig.\ref{fig:reconstruction}.

As is shown in the figure, our representation gives much more accurate reconstruction results, with clearer boundaries and richer details (like the serifs), sometimes even hard to be distinguished from the real ones. This results from our glyph construction process, where glyph shapes are combinations of primitives, which are formed by quadratic curves. This highly resembles real-world font formats, where glyphs are made of straight lines and quadratic B\'ezier curves. VAE~\cite{vae14} uses a decoder based on convolutional neural networks, which lacks structural information and infers pixel values directly, producing blurry results. Moreover, our representation well generalizes to out-of-distribution data (like the ``D" and ``C" in the figure), where VAE degrades significantly. 

\subsubsection{Glyph Interpolation}
\begin{figure}[htbp]
    \centering
    \includegraphics[width=\linewidth]{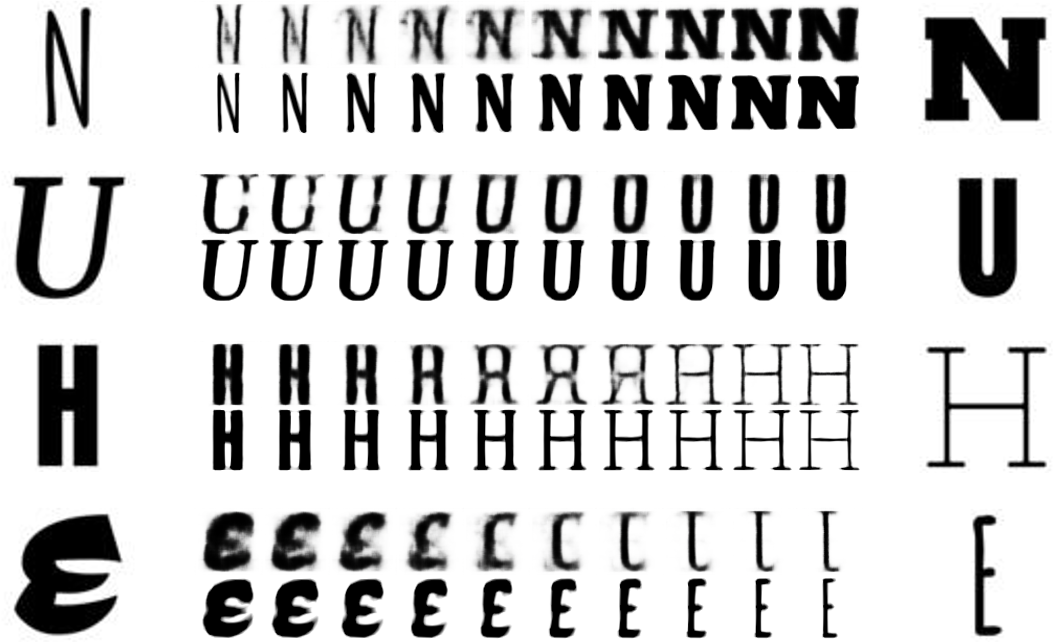}
    \caption{Visual comparisons on glyph interpolation. Large images on both sides are input glyphs of different styles. Odd rows are results of VAE and even rows are results of our model. We achieve smooth transitions between styles, benefit from the structured representation. }
    \label{fig:interpolation}
\end{figure}

\begin{figure}[htbp]
    \centering
    \includegraphics[width=\linewidth]{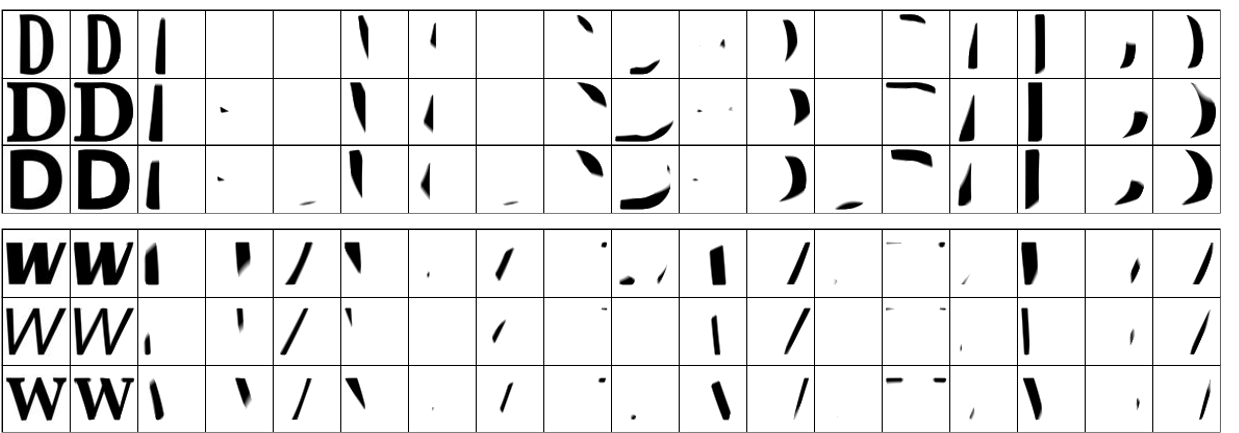}
    \caption{For each row, from left to right are the input image, the reconstruction result, and the 16 primitives learned by our representation. The primitives build correspondences among glyphs that represent the same character but have different styles, making our representation explainable and contributing to sampling qualities.}
    \label{fig:convex}
\end{figure} 

\begin{figure*}[htbp]
    \centering
    \includegraphics[width=\linewidth]{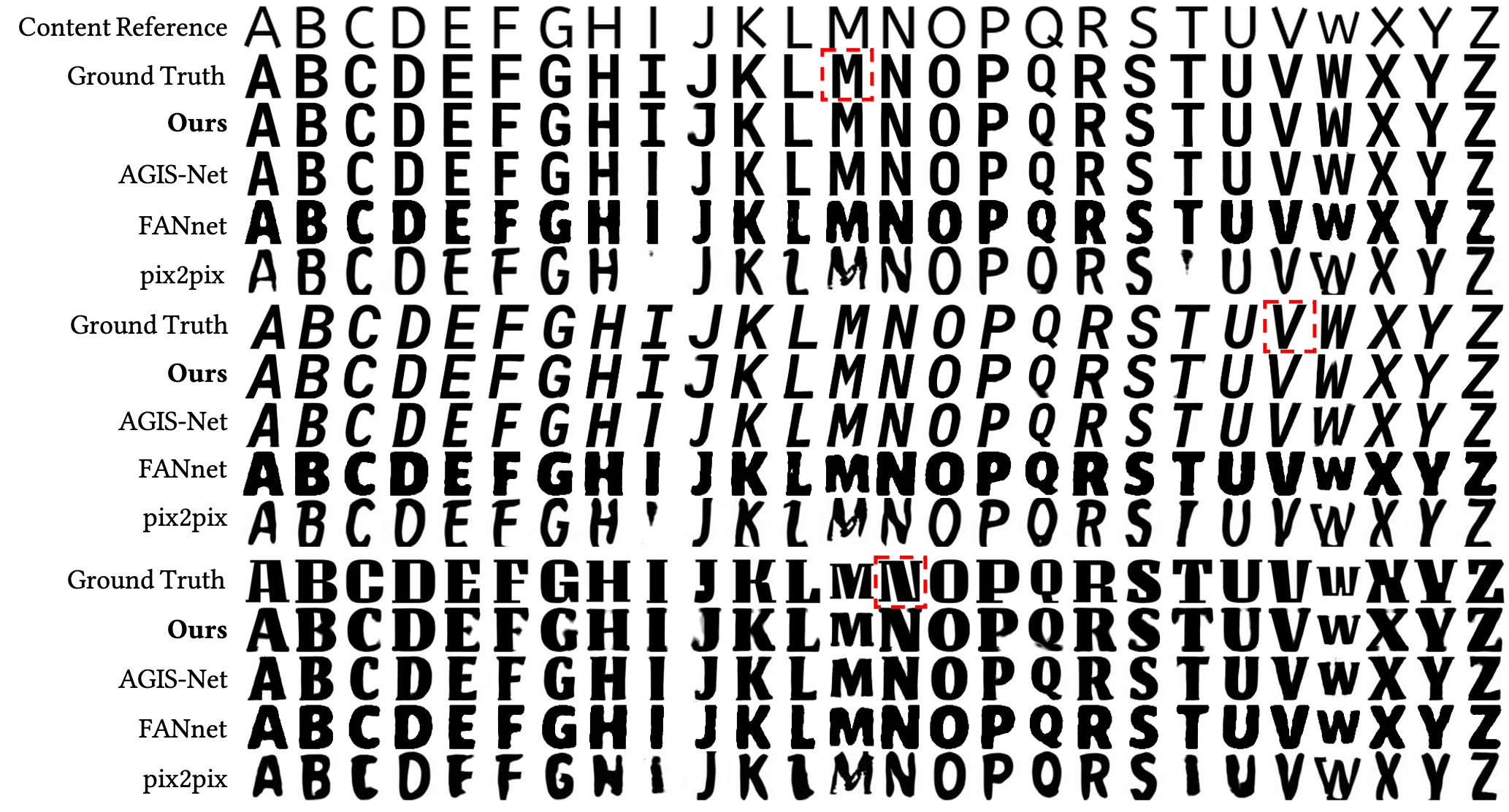}
    \caption{Visual comparisons on the one-shot font style transfer task. Content references are shown in the first row, and the style reference of each example is highlighted with red box. Our results are style-coherent, preserve more characteristics of the style references, and are most similar to the ground-truth glyphs, comparing to AGIS-Net~\cite{agisnet}, FANnet~\cite{stefann} and pix2pix~\cite{pix2pix}.}
    \label{visual-comparison}
\end{figure*}
\begin{figure*}[htbp]
    \centering
    \includegraphics[width=\linewidth]{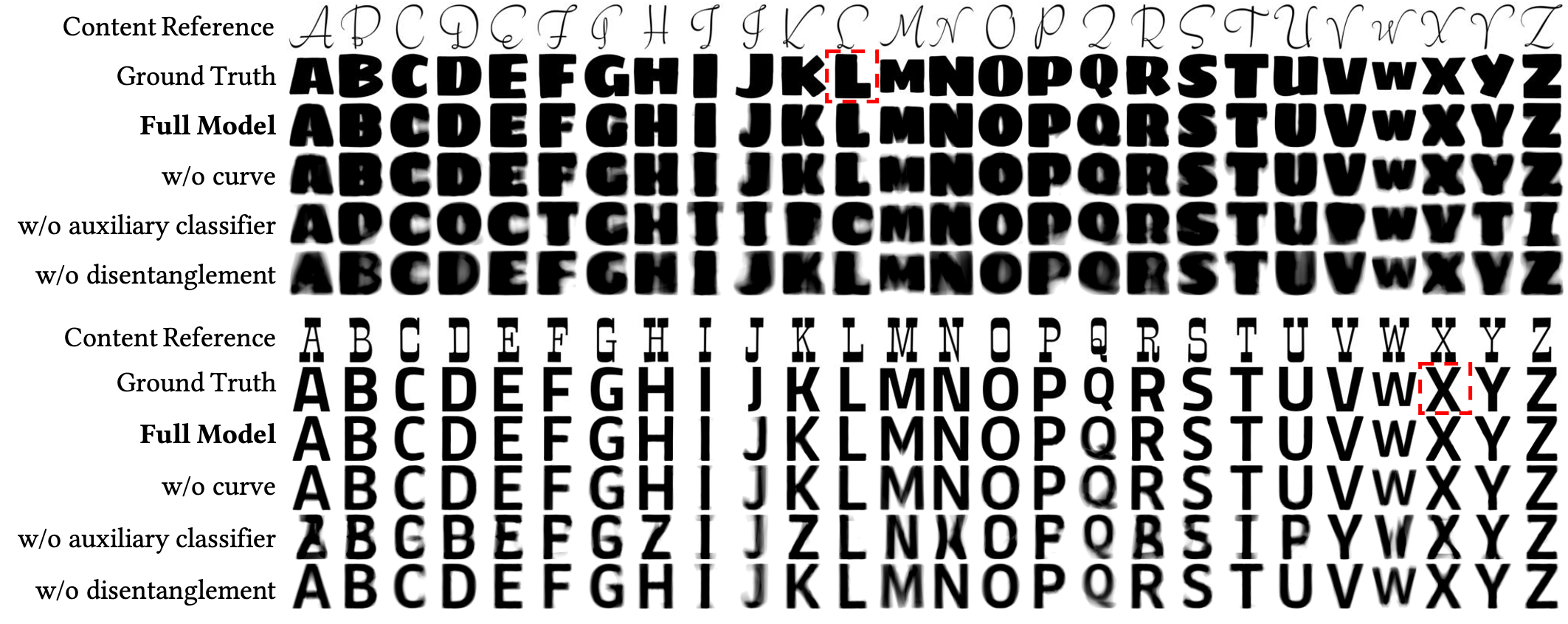}
    \caption{Qualitative results of ablation studies. For each example, content references are shown in the first row and the style reference is highlighted with red box. Our method generates the best results with all the three design factors enabled.}
    \label{fig:ablation}
\end{figure*}

Experiments on glyph interpolation further highlight the advantages of our representation. We use the same network architecture as in Sec.\ref{sec:reconstruction}. Given two glyph images $g_1, g_2$ as input, we get the corresponding latent feature vectors $z_1, z_2$ from the image encoder $\mathcal{E}_I$, and get the ``interpolated" glyph from a linear interpolation of the two vectors, whose style lies between the two glyphs. Qualitative results are shown in Fig.\ref{fig:interpolation}, where odd rows are results from VAE and evens rows are from our representation. We observe smooth style transitions in our results, and the quality of interpolated glyphs are far better than those from VAE. From the visual comparisons, we argue that our representation leads to a better latent manifold of glyphs, thus behaving well in sampling new font styles.

We believe the superiority of our representation in glyph interpolation largely contributes from our structured design. Our representation is able to automatically learn the corresponding parts of the same character in different styles, without any explicit part supervision. We visualize the primitives learned by our representation of some sampled glyphs in Fig.\ref{fig:convex}. A primitive of a glyph well corresponds to that of another glyph in the same character but different style. During interpolation, the appearance of primitives changes smoothly. This structured representation reduces learning space and helps produce more realistic results.

\subsection{One-Shot Font Style Transfer}
We compare the quality of one-shot font style transfer results generated by our model and other competitors, including pix2pix, FANnet and a one-shot adaptation of AGIS-Net. All the methods are trained and evaluated on the same dataset. We present the visual comparison results and quantitative results in Fig.\ref{visual-comparison} and Tab.\ref{tab:evaluation}.

\begin{table}%
\caption{Quantitative comparisons on the one-shot font style transfer task. Our model using the implicit representation achieves state-of-the-art results with even fewer parameters.}
\label{tab:evaluation}
\begin{minipage}{\columnwidth}
\begin{center}
\begin{tabular}{rrrrr}
  \toprule
  Method & \#Param & SSIM & L1 &  LPIPS  \\
  \midrule
  \textbf{Ours} & \textbf{1.95M} & \textbf{0.6518} & \textbf{21.62} & 0.094 \\ 
  AGIS-Net & 19.56M & 0.6066 &  23.96  & \textbf{0.076} \\
  FANnet & 4.22M & 0.3661 &  44.17     & 0.196 \\
  pix2pix  & 54.41M & 0.5192 &  49.19  & 0.269 \\ 
  \bottomrule
\end{tabular}
\end{center}
\bigskip\centering
\end{minipage}
\end{table}

Quantitative results in Tab.\ref{tab:evaluation} show that our method achieves better results on $L_1$ and SSIM, and perform closely with AGIS-Net on LPIPS, but with much fewer parameters. As is shown in Figure~\ref{visual-comparison}, our method visually outperforms all other methods under the one-shot setting, well preserving style consistency among characters, and producing promising overall shapes in terms of thickness, slopes and serifs. In comparison, pix2pix struggles to generate complete shapes. FANnet tends to produce thicker glyphs, and cannot guarantee smooth boundaries. AGIS-Net~\cite{agisnet} is the best performing method among the three baselines. It generates reasonable results with consistent styles and clear boundaries, but still fails to preserve some key characteristics of the style reference, such as width (examples on the top) and serifs (examples on the bottom). Note that our methods can extend naturally to few-shot settings, by taking the average style features of multiple style references.

\subsection{Ablation Studies}
We perform ablation studies to validate the effectiveness of several design choices of our method. We study the performance differences on the one-shot font style transfer task with or without these factors, which are listed as follows:
\begin{itemize}
    \item Whether to use quadratic curves or straight lines for primitive construction. (w or w/o curves)
    \item Whether to use the auxiliary character classifier $\mathcal{C}$. (w or w/o auxiliary classifier)
    \item Whether to apply disentanglement on style and content. If not, we simply stack the style and content reference image as input to the encoder. (w or w/o disentanglement)
\end{itemize}
Figure~\ref{fig:ablation} shows qualitative generation results with respect to different settings. With content references of uncommon styles as input (``Content Reference" row), we can see that training without auxiliary classifier fails to generate the desirable character. For example, the generated ``E" and ``L" in the first example are both like ``C" because they are easily confused with each other in the content reference font. Same for ``I", ``T" and ``F", ``P" in the second example. For the representation using straight lines instead of quadratic curves, we can observe artifacts in areas with large curvature. It is hard for primitives enclosed by straight lines to perfectly fit a curved contour, like ``J" and ``Q" in both examples. For the network without disentanglement, blurry results are generated, failing to combine the style and content information very well. We also present SSIM scores of different settings in Tab.\ref{tab:ablation} for a quantitative evaluation, which yields the same conclusion that these design choices all contribute to the final results. 

\begin{table}%
\caption{Quantitative comparison of ablation studies. All the three design choices are essential to producing satisfactory results.}
\label{tab:ablation}
\begin{minipage}{\columnwidth}
\begin{center}
\begin{tabular}{rrrrr}
  \toprule
  Model & SSIM   \\
  \midrule
  w/o disentanglement  & 0.5829 \\
  w/o auxiliary classifier & 0.6041 \\
  w/o curves &   0.6264  \\
  \textbf{Full Model} &  \textbf{0.6518} \\ 
  \bottomrule
\end{tabular}
\end{center}
\bigskip\centering
\end{minipage}
\end{table}

\subsection{Failure Cases and Future Work}
\begin{figure}[htbp]
    \centering
    \includegraphics[width=0.8\linewidth]{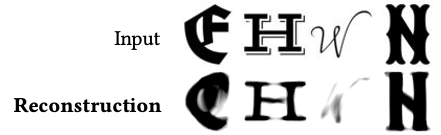}
    \caption{Failure cases on the reconstruction task. From left to right, the letters are ``E", ``H", ``W" and ``N". Our representation sometimes struggles to grasp thin structures, twisted lines and decorative elements.}
    \label{fig:failure-case}
\end{figure} 

Although our representation enables high-quality glyph modeling under most circumstances, it still struggles on some hard cases. Figure~\ref{fig:failure-case} shows some failure cases in glyph reconstruction. Our representation performs poorly on structures with extremely thin and twisted lines, like the ``W" in the figure. We suspect the reason is that thin twisted lines are difficult to model by quadratic curves. Our representation also faces degradation on glyphs with decorative elements, like the ``E" and ``N" in this figure. We believe this is because decorative elements require much more curves to be modeled. In our future work, we plan to use a multi-scale modeling procedure to fit the shape from coarse to fine, aiming to improve reconstruction quality for subtle structures, and adopt higher-order curves to enhance the representational ability.

\section{Conclusions}
\label{conclusion}
In this paper, we present a novel implicit glyph shape representation, which models glyphs as shape primitives enclosed by quadratic curves, and naturally enables generating glyph images at arbitrary high resolutions. Experiments on font reconstruction and interpolation tasks verified that this structured implicit representation is suitable for describing both structure and style features of glyphs. Furthermore, based on the proposed representation, we design a simple yet effective disentangled network for the challenging one-shot font style transfer problem, and achieve the best results comparing to state-of-the-art alternatives in both quantitative and qualitative comparisons. Benefit from this representation, our generated glyphs have the potential to be converted to vector fonts through post-processing, reducing the gap between rasterized images and vector graphics. We hope this work can provide a powerful tool for 2D shape analysis and synthesis, and inspire further exploitation in implicit representations for 2D shape modeling.

\bibliographystyle{ACM-Reference-Format}
\bibliography{sample-bibliography}

\end{document}